\pdfoutput=1

\documentclass[11pt]{article}

\usepackage{EMNLP2022}

\usepackage{times}
\usepackage{latexsym}
\usepackage{mathtools}
\usepackage{amssymb}
\usepackage{hyperref}
\usepackage{ulem}
\usepackage{stfloats}

\usepackage[T1]{fontenc}

\usepackage{verbatim}

\usepackage{makecell}
\usepackage{float}
\usepackage{algorithm}
\usepackage{algorithmic}

\usepackage[utf8]{inputenc}

\usepackage{microtype}

\usepackage{inconsolata}

\title{PATS: Sensitivity-aware Noisy Learning for Pretrained Language Models}

\author{Yupeng Zhang\textsuperscript{1}\thanks{\ \ Work done during internship at Meituan Inc. The first two authors have equal contributions.}, Hongzhi Zhang\textsuperscript{2}\footnotemark[1], Sirui Wang\textsuperscript{2}, Wei Wu\textsuperscript{2} \and Zhoujun Li\textsuperscript{1}\thanks{ \ \ Corresponding author.}\\
$^1$Beihang University, Beijing, China \quad $^2$Meituan Inc., Beijing, China \\ 
\texttt{\{G0vi\_qyx, lizj\}@buaa.edu.cn} \\
\texttt{\{zhanghongzhi03, wangsirui, wuwei30\}@meituan.com}
}

\begin{document}
\maketitle
\begin{abstract}
A wide range of NLP tasks benefit from the fine-tuning of pretrained language models (PLMs).
However, a number of redundant parameters which contribute less to the downstream task are observed in a directly fine-tuned model. 
We think the gap between pretraining and downstream tasks hinders the training of these redundant parameters, and results in a suboptimal performance of the overall model.
In this paper, we present PATS (\textbf{P}erturbation \textbf{A}ccording \textbf{T}o \textbf{S}ensitivity), a noisy training mechanism which considers each parameter's importance in the downstream task to help fine-tune PLMs. 
The main idea of PATS is to add bigger noise to parameters with lower sensitivity and vice versa, in order to activate more parameters' contributions to downstream tasks without affecting the sensitive ones much. 
%
Extensive experiments conducted on different tasks of the GLUE benchmark show PATS can consistently empower the fine-tuning of different sizes of PLMs, and the parameters in the well-performing models always have more concentrated distributions of sensitivities, which experimentally proves the effectiveness of our method.
\end{abstract}

\section{Introduction}

With a huge number of model parameters and well designed training objectives, pretrained language models (PLMs) have brought a new era to NLP \citep{bertnb1,bertnb2,bertnb3,bertnb4}. Fine-tuning PLMs such as BERT \citep{bert} has become a basic and effective way in many downstream tasks \citep{finetune1, finetune2, finetune3}. 

However, recent study has shown that aggressive fine-tuning can induce an unstable and suboptimal performance of the models especially with insufficient data \citep{bad1, bad2}, which attracts some researchers to figure out the culprits and explore effective methods to solve them \citep{bettertune1, bettertune2, bettertune3}. For example, there are some regularization methods like RecAdam \citep{Recadam} and Mixout \citep{mixout}, and adversarial training techniques like SMART \citep{smart} and FreeLB \citep{FreeLB} to alleviate the overfitting of data in downstream tasks;
Beyond that, \citet{noisytune} proposed NoisyTune with the argument that in addition to the overfitting of the limited downstream data, there could also exist overfitting in pretraining tasks, which could result in enormous gaps between pretraining and downstream task data. 
In order to overcome the gaps, NoisyTune simply adds 
some noise to parameters in the PLM before fine-tuning.
Besides, it has also been demonstrated that the existence of a large number of redundant parameters could also be a factor in the suboptimal performances of aggressively fine-tuned PLMs \citep{redundancy1, redundancy2, redundancy3}.
Considering the redundant parameters in a model are not insufficiently trained, \citet{sage} proposed a learning rate scheduler named SAGE in which larger learning rates are assigned to these parameters of low sensitivity (a measure of parameter's importance to downstream tasks). 

\begin{figure}
    \centering
    \includegraphics[width=\linewidth,scale=1]{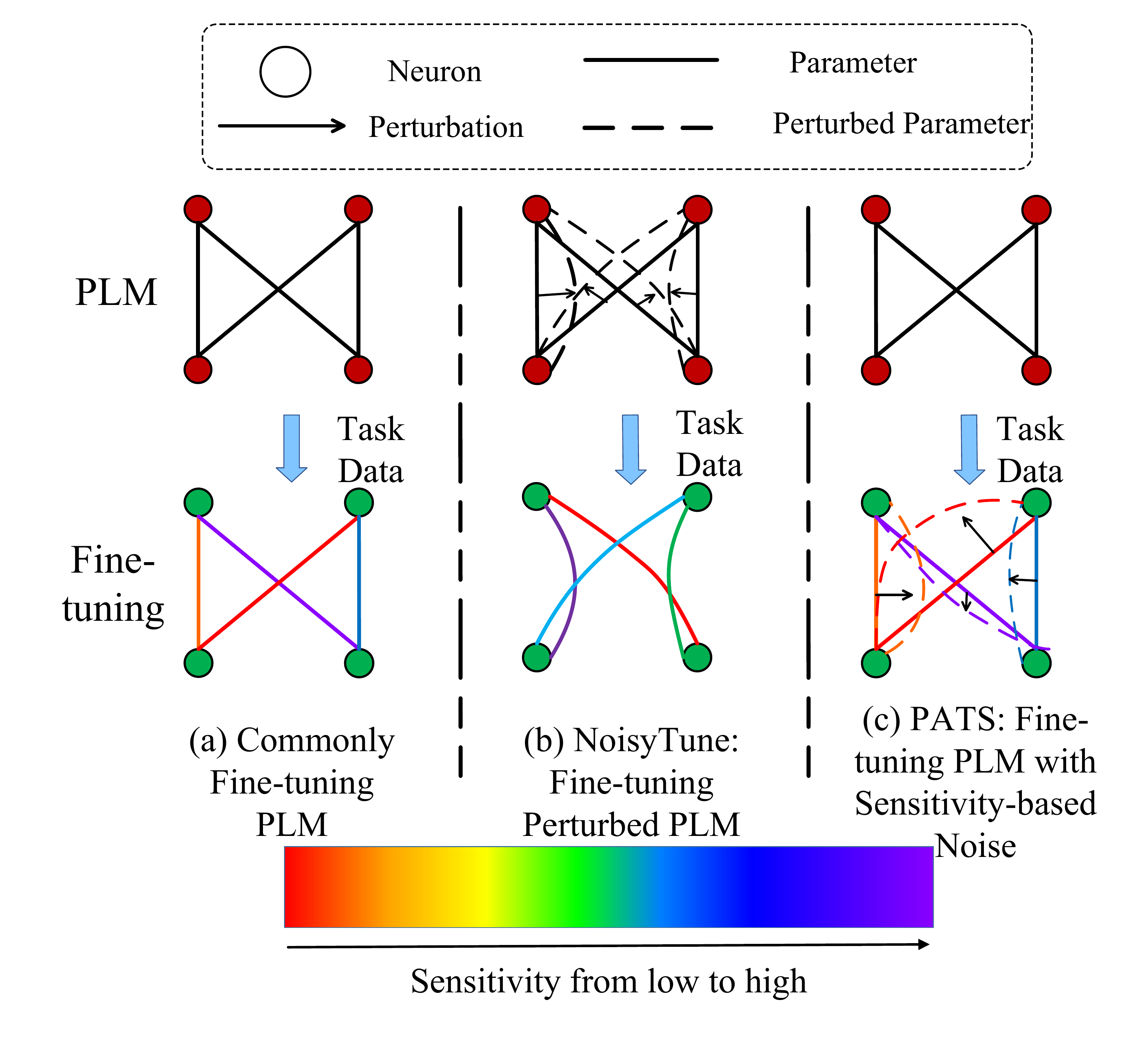}
    \caption{Different schemata of fine-tuning PLMs. \\In NoisyTune, a matrix of parameters in a PLM are perturbed with the same intensity before fine-tuning;\\
    In PATS, parameters with lower sensitivity (the "red" parameter) to downstream data are added with larger noise, and vice versa (like the "purple" parameter).}
    \label{fig:methods}
\end{figure}

There could be some connection between the gaps caused by overfitting of pretraining tasks and the redundancy of parameters. 
We consider it could be the gaps between pretraining and downstream tasks that hinder the training of these redundant parameters. 
SAGE enlarges the learning rates of insensitive parameters to help their training. However, with the sensitivity measurement considered, the insensitive parameters usually have smaller gradients, so enlarged learning rates may help them little to escape the sub-optimal areas compared to involving additional noise. 
One noisy training method to alleviate the gaps is NoisyTune, in which parameters of a matrix in a PLM are added with noise according to the standard deviation of the matrix before fine-tuning.
Nevertheless, there are few explanations about why or whether the parameters in the same matrix should be perturbed with the same intensity. Considering different parameters have different contributions to the model, noise from a unified distribution may disturb knowledge of some sensitive parameters, resulting in a loss of performance. Besides,
since each task needs to capture an appropriate textual pattern and the data of it usually comes from a special domain,
different downstream tasks could have different kinds of gaps with those of the pretraining. So the noise added to overcome the gaps should also be related to the downstream task data.
In this paper, we propose a novel parameter-wise noisy fine-tuning method called PATS (\textbf{P}erturbation \textbf{A}ccording \textbf{T}o \textbf{S}ensitivity) to make full use of perturbation on parameters to handle the problems above.
We focus on balancing the contributions of all parameters in the model by activating the insensitive ones to play better roles in downstream tasks. So the main idea of our method is adding different intensities of noise to parameters according to their sensitivity when fine-tuning PLMs, different from NoisyTune (Fig. \ref{fig:methods} (b)) in which noise added to a matrix of parameters is from a unified distribution and unrelated to downstream task data. Specifically, during fine-tuning in PATS (Fig. \ref{fig:methods} (c)),
larger noise will be added to the parameters with lower sensitivity (such as the parameter shown in red), while sensitive parameters (such as the parameter shown in purple) will be barely perturbed. 
Our contributions can be summarized as follows: 1) We propose a simple but effective method to help all parameters be trained sufficiently when fine-tuning PLMs in downstream tasks. 2) Among all the training methods with noise, PATS is the first sensitivity-aware one which perturbs models with noise of different distributions according to parameters' sensitivity, to the best of our knowledge. 3) Extensive experiments on the GLUE benchmark show PATS makes a difference in boosting the performance of PLMs in downstream NLP tasks.

\section{Approach}
In this section, we present our PATS for PLMs fine-tuning. 
Previous matrix-wise noisy methods perturb a PLM by adding noise from a uniform distribution to a matrix of parameters.
Different from them, in PATS, each parameter even from the same matrix will be paid to different attention according to its sensitivity. 
It is also worth noting that in PATS, a PLM is not perturbed in advance like NoisyTune, instead the perturbation happens during training as the task data comes in. 
In the following sections, we will introduce the calculation of parameter sensitivity first and then present the noisy learning mechanism in detail.

\subsection{Sensitivity Measurement}

The sensitivity of a parameter is used to measure the change of the output or loss after setting it to zero \citep{sensitivity3, sensitivity4, sensitivity5, sensitivity6, sensitivity7}. To be specific, given a BERT-like pre-trained language model \textbf{M} with parameters $\mathbf{\Theta}=\{\theta_1, \theta_2,\cdots, \theta_n\}\in  \mathbb{R}^n$, the sensitivity of the $j$-th parameter $\theta_j$ is written as $s_j$, which can be defined as:

\begin{align}
s_j &=|\mathcal{L}(\mathbf{\Theta})-\mathcal{L}(\theta_1, \cdots, \theta_{j-1}, 0, \theta_{j+1}, \cdots, \theta_n)| \nonumber \\
&\approx |\theta_j\nabla_{\theta_j}\mathcal{L}(\mathbf{\Theta})|,
\end{align}
where $\mathcal{L}$ is a loss function and we use the first-degree Taylor polynomial to approximate $s_j$ ignoring the higher order remainder to accelerate the calculation of it.

In order to avoid huge oscillation of $s_j$ caused by an abnormal batch of data, we adopt the exponential moving average of $s_j$ used in many other models and optimizers \citep{sage,exp-average1, exp-average2, exp-average3} as the real sensitivity indicator, which can be expressed by the following equation:
\begin{align}
    \bar{s_j}=\beta \bar{s_j}^*+(1-\beta)s_j, \beta\in (0, 1),
\end{align}
where $\bar{s_j}$ and $\bar{s_j}^*$ are the exponential moving average of $s_j$ in the current and previous iteration. $\beta$ is a hyper-parameter used to adjust the importance of $s_j$ calculated by the current batch of data. 

\subsection{Training with Noise}

\label{ap:algo}
\begin{algorithm}
\caption{PATS for Adamax(max($\cdot$) returns a matrix with the maximum values of each element of the input matrices or vectors; sum($\cdot$) returns a scalar equal to the sum of all values of a matrix or vector; $\mathbf{I}$ denotes an all-ones matrix; $\odot$ denotes Hadamard product and $\oslash$ denotes Hadamard division)}\label{al:useperturb}
\textbf{Input:}Step size $\alpha$; Model parameters $\mathbf{\Theta}\in \mathbb{R}^n$; Number of training iterations $T$; Number of parameters in the current matrix $N$; Exponential decay rates $\beta, \beta_1, \beta_2\in [0,1)$; Basic noise $\lambda$; Minimum effective sensitivity indicator $\gamma$; A small number that prevents an error of dividing by zero $\epsilon\in(0, 1)$; Data $\mathcal{D}$; Loss function $\mathcal{L}(\cdot)$.
\begin{algorithmic}[1]
\STATE Initialize $\mathbf{\widetilde{S}}^{(0)}\leftarrow \mathbf{0}\in \mathbb{R}^{N}$.
\STATE Initialize $\mathbf{M}^{(0)}\leftarrow \mathbf{0}\in \mathbb{R}^{N}$.
\STATE Initialize $\mathbf{U}^{(0)}\leftarrow \mathbf{0}\in \mathbb{R}^{N}$.
\FOR{$t\leftarrow1$ to $T$}
\STATE $\mathbf{d}^{(t)}\stackrel{sample}{\longleftarrow}\mathcal{D}$.
\STATE $\mathbf{G}^{(t)}\leftarrow\nabla_{\mathbf{\Theta}^{(t)}}\mathcal{L}(\mathbf{d}^{(t)},\mathbf{\Theta}^{(t)})$.
\STATE $\mathbf{S}^{(t)}\leftarrow\mathbf{\Theta}^{(t)}\odot \mathbf{G}^{(t)}$.
\STATE $\mathbf{M}^{(t)}\leftarrow\beta_1 \mathbf{M}^{(t-1)}+(1-\beta_1)\mathbf{G}^{(t)}$.
\STATE $\mathbf{U}^{(t)}\leftarrow\mbox{max}(\beta_2\mathbf{U}^{(t-1)},|\mathbf{G}^{(t)}|)$.
\STATE $\mathbf{\widetilde{S}}^{(t)}\leftarrow\beta\mathbf{\widetilde{S}}^{(t-1)}+(1-\beta)\mathbf{S}^{(t)}$.
\STATE $\mathbf{R}\leftarrow\lambda\mbox{max}(\mbox{sum}(\mathbf{\widetilde{S}}^{(t)})\textbf{I}\oslash(N\mathbf{\widetilde{S}}^{(t)}+\epsilon\mathbf{I})-\gamma \mathbf{I},\mathbf{0})$.
\STATE $\mathbf{Q}\sim \mathcal{N}(\mathbf{0}, \mathbf{R})$.
\STATE $\mathbf{Z}\sim \mathcal{B}(N,p)$.
\STATE $\mathbf{\Theta}^{(t+1)}\leftarrow\mathbf{\Theta}^{(t)}-(\alpha/(1-\beta_1^t))\mathbf{M}^{(t)}\oslash \mathbf{U}^{(t)}+\mathbf{Q}\odot \mathbf{Z}$.
\STATE $t\leftarrow t+1$
\ENDFOR

\end{algorithmic}
\end{algorithm}

Our goal is to mainly activate the contributions of less sensitive parameters by perturbing them with bigger noise and leave parameters with larger sensitivity less affected at the same time. In our framework, we use a hyper-parameter $\lambda$ as initial noise and the degree of perturbation to different parameters will be scaled up and down based on it according to their sensitivity. The intensity of perturbation can be formulated by the following equations:

\begin{gather}
    \bar{s}=\dfrac{1}{N}\sum_{i=1}^N\bar{s_i} \label{eq:ave}\\
    r_j=\lambda\cdot\max(\dfrac{\bar{s}}{\bar{s_j}+\epsilon}-\gamma,0),0<\epsilon\ll 1 \label{eq:scale}
\end{gather}

In Eq. \ref{eq:ave}, $\bar{s}$ is the average sensitivity of the matrix containing $\theta_j$ with $N$ parameters. $r_j$ in Eq. \ref{eq:scale} means the intensity of the noise to be added on a parameter $\theta_j$, which is scaled on $\lambda$ by the division of $\bar{s}$ and $\bar{s_j}$. $\epsilon$ is a small number used to prevent zero denominator. Since $r_j$ and $\bar{s_j}$ are inversely correlated, the intensity of noise added to every parameter with a lower sensitivity than the average will be larger than $\lambda$, and vice versa
as we expect. As for the reason why $\bar{s}$ is restricted to the current matrix, as we found, the value distributions of different matrix parameters are sometimes very different. For example, values of parameters in matrix $\mathbf{A}$ are significant higher than those in matrix $\mathbf{B}$. And with the sensitivity measurement considered, sensitive parameters are usually themselves large on value. So if $\bar{s}$ is calculated based on all parameters of the model, some matrices of parameters with low values and sensitivity may be perturbed fiercely to some values very far from their original ones, which has unstable performances on experiments.
To further reduce the perturbation on sensitive parameters and let them keep regular gradient-driven update, we use a margin constant $\gamma$ in Eq. \ref{eq:scale} to zero-out the noise added on the parameters that are highly sensitive. 

\begin{table*}[ht]
\centering
\resizebox{\textwidth}{!}{
\begin{tabular}{lccccccccc}
\Xhline{1.6pt}
\textbf{Model}  & \makecell[c]{\textbf{CoLA}\\Mcc} & \makecell[c]{\textbf{MRPC}\\F1} & \makecell[c]{\textbf{RTE}\\Acc} & \makecell[c]{\textbf{STS-B}\\Pcc}  & \makecell[c]{\textbf{QQP}\\F1} & \makecell[c]{\textbf{QNLI}\\Acc} & \makecell[c]{\textbf{MNLI}\\Acc} & \makecell[c]{\textbf{SST}\\Acc} & \makecell[c]{\textbf{Avg}\\Score}\\
\hline

$\mbox{BERT}_{\mbox{base}}$  & 58.94  & 90.19  & 68.03 & 89.28 & 88.53 & 91.96 & 84.63 & 92.77 & 83.18\\

$\mbox{BERT}_{\mbox{base}}+\mbox{SAGE}$  & 59.45  & 90.53  & 71.78 & 89.81 & 88.61 & 91.87 & 84.59 & \textbf{93.06} & 83.65\\

$\mbox{BERT}_{\mbox{base}}+\mbox{NoisyTune}$ & 60.01  & 90.34  & 69.71 & 89.81 & 88.58 & 91.82 & 84.64 & 92.85 & 83.47\\

$\mbox{BERT}_{\mbox{base}}+\mbox{PATS}$  & \textbf{60.67} & \textbf{91.05} & \textbf{72.08} & \textbf{89.86} & \textbf{88.64} & \textbf{92.02} & \textbf{84.80} & 92.89 & \textbf{84.00}\\
\hline
$\mbox{RoBERTa}_{\mbox{large}}$  & 66.67  & 91.89  & 85.44 & 91.98 & 89.26 & 94.45 & 90.25 & 96.10 & 88.25\\

$\mbox{RoBERTa}_{\mbox{large}}+\mbox{SAGE}$  & 67.36  & 93.27  & 85.56 & 92.05 & 89.27 & 94.54 & 90.25 & 96.25 & 88.57\\

$\mbox{RoBERTa}_{\mbox{large}}+\mbox{NoisyTune}$ & 67.47  & 93.26  & 85.52 & 92.00 & 89.36 & 94.53 & 90.04 & 96.12 & 88.56\\

$\mbox{RoBERTa}_{\mbox{large}}+\mbox{PATS}$ & \textbf{68.62}  & \textbf{93.52} & \textbf{86.29} & \textbf{92.23} & \textbf{89.40} & \textbf{94.64} & \textbf{90.44} & \textbf{96.30} & \textbf{88.90}\\
\Xhline{1.6pt}
\end{tabular}
}

\caption{Results of models on GLUE dev set. }
\label{tab:GLUE}
\end{table*}

For each parameter $\theta_j$, the noise $q_j$ that may finally be added to it is independently randomly sampled from a Gaussian distribution with the mean of zero and the standard deviation of $\sigma_j$ as $q_j\sim N(0, \sigma_j^2)$, where $\sigma_j=\sqrt{r_j}$. So in an iteration, we update each parameter by:
\begin{align}
    \widetilde{\theta_j}=\theta_j-\eta\cdot\nabla_{\theta_j}\mathcal{L}(\mathbf{\Theta}) + q_j\cdot z \label{eq:update},
\end{align}
where $\eta$ is learning rate and $z\sim B(1, p)$ is a random value sampled from Bernoulli distribution which outputs $1$ with probability $p$ and $0$ with probability $1-p$. Algorithm \ref{al:useperturb} 
shows the PATS algorithm for Adamax \citep{Adamax} optimizer.

\section{Experiments}

\subsection{Datasets and Baselines}
We conduct extensive experiments on the eight tasks of the GLUE benchmark \citep{glue} and adopt the publicly available BERT-base \citep{bert} and RoBERTa-large \citep{RoBERTa} models on every task individually. The following three baselines are selected for comparison: (1) \textbf{Standard PLM fine-tuning}, which fine-tunes PLMs directly; (2) \textbf{NoisyTune} \citep{noisytune}, which is a noisy training method that adds matrix-wise noise before fine-tuning; (3) \textbf{SAGE} \citep{sage}, which is an optimized learning rate schedule which adjusts the learning rate of every parameter according to its sensitivity.

\subsection{Performance Evaluation}

On each task, we repeat our experiments 5 times with different random seeds and report the average scores of every model, which are shown in Table \ref{tab:GLUE}.\footnote{The results of the MNLI task are obtained by averaging the output accuracies of the models on the mnli-matched dataset and the mnli-mismatched dataset.}
According to the results, PATS optimized models consistently outperforms directly fine-tuned ones on different downstream tasks, especially on those with small datasets (CoLA \& MRPC \& RTE).
Specifically, PATS improves by around 2 points on CoLA and RTE, and around 1 point on MRPC. 
In addition, as a parameter-wise method based on sensitivity, PATS experimentally outperforms the matrix-wise noisy method NoisyTune and the sensitivity-based learning rate scheduler SAGE on 7 out of the 8 tasks.
The experimental results demonstrate the effectiveness of PATS. 

\subsection{Empirical Analysis}
In this section, we conduct additional analyses on sensitivity of parameters in the fine-tuned models.
\begin{figure}
    \centering
    \includegraphics[width=\linewidth,scale=10]{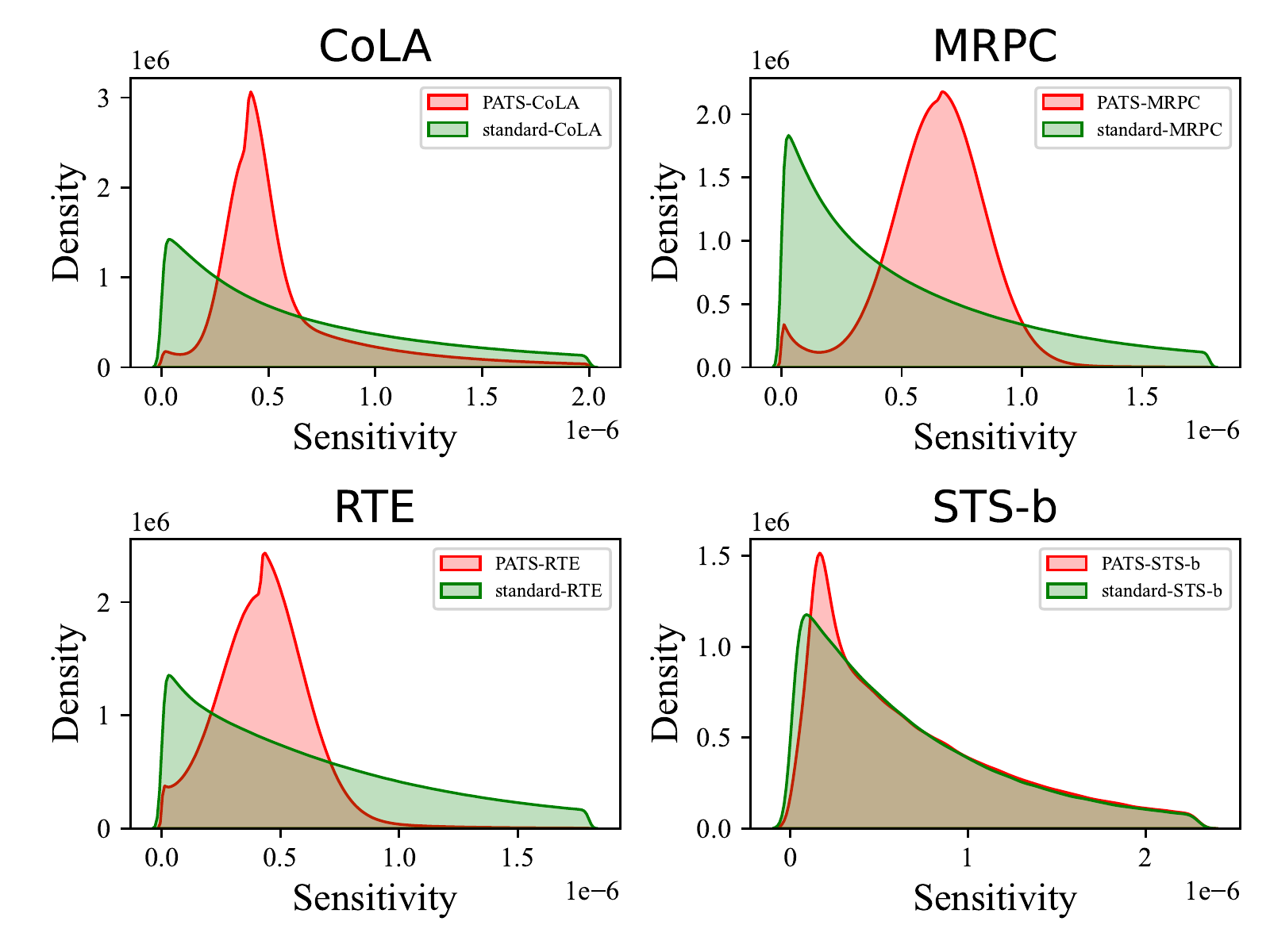}
    \caption{The sensitivity distributions of the parameters in different optimized models.}
    \label{fig:distribution}
\end{figure}

Fig. \ref{fig:distribution} shows the sensitivity distribution of the model parameters fine-tuned in different ways.
It is found that the sensitivity of the parameters in the PATS optimized models is more tightly clustered than that in the models fine-tuned in the common way. Besides, there remain fewer insensitive parameters in PATS optimized models than those in baseline models. And it is no longer a few high-sensitive parameters that dominate the models as what happens in normal fine-tuning, which indicates that perturbation helps parameters with low sensitivity gain more attention during training and lets the contribution of each parameter in the optimized models more balanced.

\begin{figure}
    \centering
    \includegraphics[width=\linewidth,scale=10]{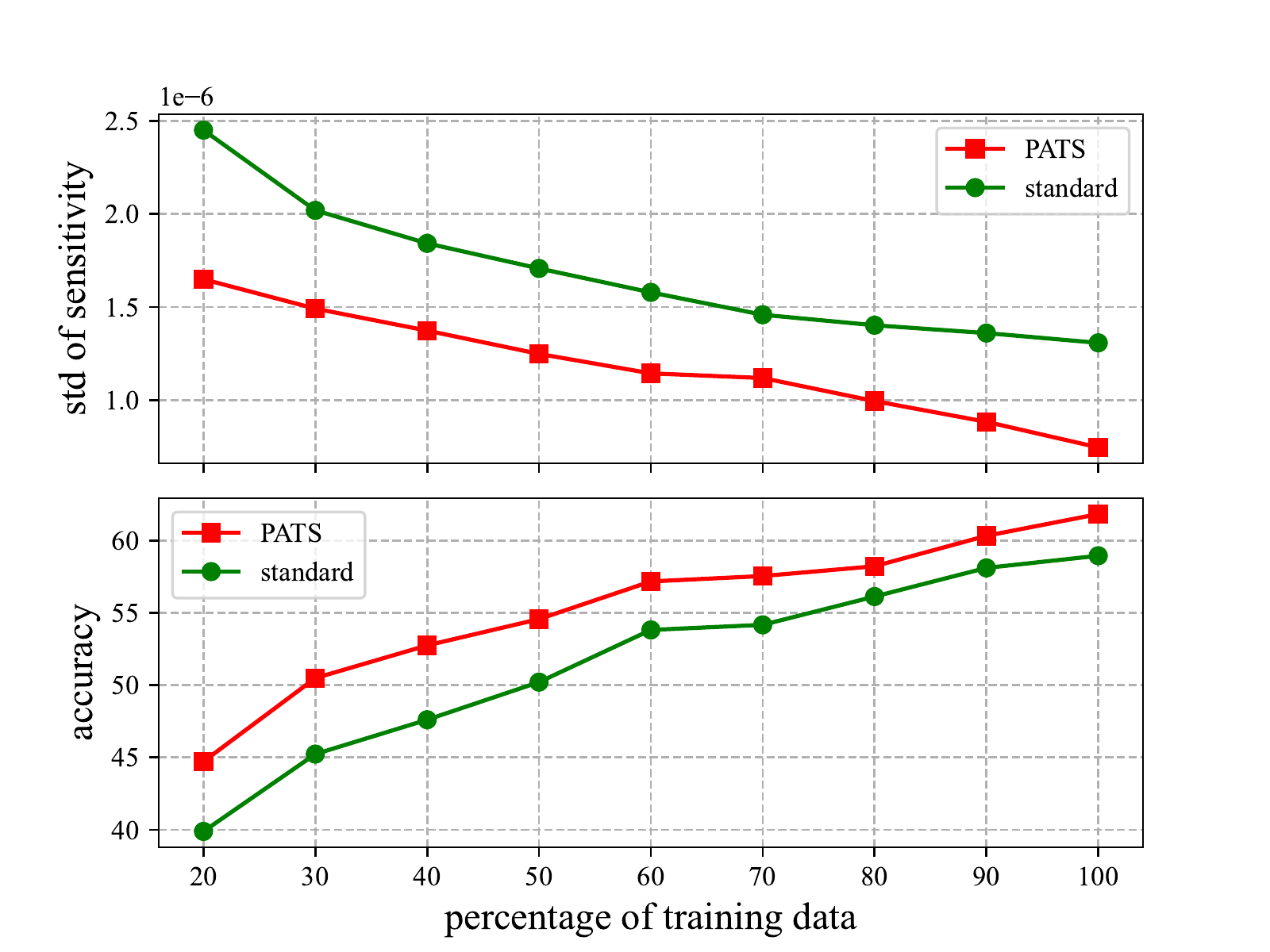}
    \caption{Performances of PATS on data of different sizes.}
    \label{fig:lines}
\end{figure}

To further investigate the effect of PATS on small datasets, we also post the accuracies of the models fine-tuned on different proportions of training data sampled from the CoLA\footnote{The phenomena observed on other tasks are similar.} dataset with and without PATS.
Fig. \ref{fig:lines} shows PATS optimized models consistently outperform directly fine-tuned ones on different sizes of datasets, demonstrating the generalizability of the approach. 
Moreover, we can also observe that as the size of training data increases, the performances of the models improve along with concomitant decreases in the standard deviations of sensitivity. This phenomenon further indicates that training with limited data will lose some of the performance capabilities of PLMs by leaving more undertrained or insensitive parameters, because small datasets is insufficient for PLMs to overcome the gap between pretraining and downstream tasks. The inverse correlation between accuracy and sensitivity concentration justifies our original intention of balancing the sensitivity of parameters. And the displayed performances experimentally demonstrate its availability.

\section{Conclusion}

We propose a novel noisy training method called PATS to optimize fine-tuning of PLMs. 
Since aggressive fine-tuning PLMs will leave a large number of insensitive parameters which contribute little to the overall model, PATS activates them and balance the contributions of all parameters in downstream tasks
by adding noise to each parameter according to its sensitivity in the process of training.
PATS is a simple mechanism without much computational and memory overhead compared to adversarial training which requires additional backwards passes. 
Extensive experiments on eight tasks of the GLUE benchmark show that PATS can consistently improve the performance of PLMs on downstream tasks with the sensitivity of the parameters more concentrated, which is especially pronounced on small datasets.

\section*{Limitations}
PATS introduces four additional hyperparameters, which increases some work of users on hyperparameter tuning. For example, a too small $\lambda$ could make few differences while an overlarge $\lambda$ may result in unstable performances of models. Though we have summarized effective parameter configurations on the NLU tasks of the GLUE benchmark, it cannot guarantee that these settings are still applicable on other tasks such as neural machine translation. We will explore the connections between the hyperparameters in theory and narrow the search ranges of the hyperparameter group in future work.

\normalem
\bibliography{anthology,custom}

\begin{thebibliography}{47}
\expandafter\ifx\csname natexlab\endcsname\relax\def\natexlab#1{#1}\fi

\bibitem[{Bao et~al.(2020)Bao, Dong, Wei, Wang, Yang, Liu, Wang, Gao, Piao,
  Zhou, and Hon}]{dev}
Hangbo Bao, Li~Dong, Furu Wei, Wenhui Wang, Nan Yang, Xiaodong Liu, Yu~Wang,
  Jianfeng Gao, Songhao Piao, Ming Zhou, and Hsiao-Wuen Hon. 2020.
\newblock \href {https://proceedings.mlr.press/v119/bao20a.html} {{U}ni{LM}v2:
  Pseudo-masked language models for unified language model pre-training}.
\newblock In \emph{Proceedings of the 37th International Conference on Machine
  Learning}, volume 119 of \emph{Proceedings of Machine Learning Research},
  pages 642--652. PMLR.

\bibitem[{Bar-Haim et~al.(2006)Bar-Haim, Dagan, Dolan, Ferro, Giampiccolo,
  Magnini, and Szpektor}]{RTE2}
Roy Bar-Haim, Ido Dagan, Bill Dolan, Lisa Ferro, Danilo Giampiccolo, Bernardo
  Magnini, and Idan Szpektor. 2006.
\newblock The second pascal recognising textual entailment challenge.
\newblock In \emph{Proceedings of the second PASCAL challenges workshop on
  recognising textual entailment}, volume~6, pages 6--4. Venice.

\bibitem[{Bentivogli et~al.(2009)Bentivogli, Clark, Dagan, and
  Giampiccolo}]{RTE4}
Luisa Bentivogli, Peter Clark, Ido Dagan, and Danilo Giampiccolo. 2009.
\newblock The fifth pascal recognizing textual entailment challenge.
\newblock In \emph{TAC}.

\bibitem[{Bowman et~al.(2015)Bowman, Angeli, Potts, and Manning}]{MNLI2}
Samuel~R Bowman, Gabor Angeli, Christopher Potts, and Christopher~D Manning.
  2015.
\newblock A large annotated corpus for learning natural language inference.
\newblock \emph{arXiv preprint arXiv:1508.05326}.

\bibitem[{Cer et~al.(2017)Cer, Diab, Agirre, Lopez-Gazpio, and Specia}]{STS-b}
Daniel Cer, Mona Diab, Eneko Agirre, I{\~n}igo Lopez-Gazpio, and Lucia Specia.
  2017.
\newblock \href {https://doi.org/10.18653/v1/S17-2001} {{S}em{E}val-2017 task
  1: Semantic textual similarity multilingual and crosslingual focused
  evaluation}.
\newblock In \emph{Proceedings of the 11th International Workshop on Semantic
  Evaluation ({S}em{E}val-2017)}, pages 1--14, Vancouver, Canada. Association
  for Computational Linguistics.

\bibitem[{Chen et~al.(2020)Chen, Hou, Cui, Che, Liu, and Yu}]{Recadam}
Sanyuan Chen, Yutai Hou, Yiming Cui, Wanxiang Che, Ting Liu, and Xiangzhan Yu.
  2020.
\newblock \href {https://doi.org/10.18653/v1/2020.emnlp-main.634} {Recall and
  learn: Fine-tuning deep pretrained language models with less forgetting}.
\newblock In \emph{Proceedings of the 2020 Conference on Empirical Methods in
  Natural Language Processing (EMNLP)}, pages 7870--7881, Online. Association
  for Computational Linguistics.

\bibitem[{Clark et~al.(2020)Clark, Luong, Le, and Manning}]{wnli2}
Kevin Clark, Minh-Thang Luong, Quoc Le, and Christopher~D. Manning. 2020.
\newblock \href {https://doi.org/10.18653/v1/2020.emnlp-main.20} {Pre-training
  transformers as energy-based cloze models}.
\newblock In \emph{Proceedings of the 2020 Conference on Empirical Methods in
  Natural Language Processing (EMNLP)}, pages 285--294, Online. Association for
  Computational Linguistics.

\bibitem[{Dagan et~al.(2005)Dagan, Glickman, and Magnini}]{RTE1}
Ido Dagan, Oren Glickman, and Bernardo Magnini. 2005.
\newblock The pascal recognising textual entailment challenge.
\newblock In \emph{Machine Learning Challenges Workshop}, pages 177--190.
  Springer.

\bibitem[{Dalvi et~al.(2020)Dalvi, Sajjad, Durrani, and Belinkov}]{redundancy3}
Fahim Dalvi, Hassan Sajjad, Nadir Durrani, and Yonatan Belinkov. 2020.
\newblock \href {https://doi.org/10.18653/v1/2020.emnlp-main.398} {Analyzing
  redundancy in pretrained transformer models}.
\newblock In \emph{Proceedings of the 2020 Conference on Empirical Methods in
  Natural Language Processing (EMNLP)}, pages 4908--4926, Online. Association
  for Computational Linguistics.

\bibitem[{Devlin et~al.(2019)Devlin, Chang, Lee, and Toutanova}]{bert}
Jacob Devlin, Ming-Wei Chang, Kenton Lee, and Kristina Toutanova. 2019.
\newblock \href {https://doi.org/10.18653/v1/N19-1423} {{BERT}: Pre-training of
  deep bidirectional transformers for language understanding}.
\newblock In \emph{Proceedings of the 2019 Conference of the North {A}merican
  Chapter of the Association for Computational Linguistics: Human Language
  Technologies, Volume 1 (Long and Short Papers)}, pages 4171--4186,
  Minneapolis, Minnesota. Association for Computational Linguistics.

\bibitem[{Ding et~al.(2019)Ding, ding, Zhou, Guo, Han, and Liu}]{sensitivity5}
Xiaohan Ding, guiguang ding, Xiangxin Zhou, Yuchen Guo, Jungong Han, and
  Ji~Liu. 2019.
\newblock \href
  {https://proceedings.neurips.cc/paper/2019/file/f34185c4ca5d58e781d4f14173d41e5d-Paper.pdf}
  {Global sparse momentum sgd for pruning very deep neural networks}.
\newblock In \emph{Advances in Neural Information Processing Systems},
  volume~32. Curran Associates, Inc.

\bibitem[{Dodge et~al.(2020)Dodge, Ilharco, Schwartz, Farhadi, Hajishirzi, and
  Smith}]{bad1}
Jesse Dodge, Gabriel Ilharco, Roy Schwartz, Ali Farhadi, Hannaneh Hajishirzi,
  and Noah Smith. 2020.
\newblock \href {https://doi.org/10.48550/ARXIV.2002.06305} {Fine-tuning
  pretrained language models: Weight initializations, data orders, and early
  stopping}.
\newblock \emph{arXiv preprint arXiv:2022.06305}.

\bibitem[{Fan et~al.(2019)Fan, Grave, and Joulin}]{redundancy1}
Angela Fan, Edouard Grave, and Armand Joulin. 2019.
\newblock \href {https://doi.org/10.48550/ARXIV.1909.11556} {Reducing
  transformer depth on demand with structured dropout}.
\newblock \emph{arXiv preprint arXiv:1909.11556}.

\bibitem[{Giampiccolo et~al.(2007)Giampiccolo, Magnini, Dagan, and
  Dolan}]{RTE3}
Danilo Giampiccolo, Bernardo Magnini, Ido Dagan, and Bill Dolan. 2007.
\newblock The third pascal recognizing textual entailment challenge.
\newblock In \emph{Proceedings of the ACL-PASCAL workshop on textual entailment
  and paraphrasing}, pages 1--9. Association for Computational Linguistics.

\bibitem[{Guu et~al.(2020)Guu, Lee, Tung, Pasupat, and Chang}]{bertnb1}
Kelvin Guu, Kenton Lee, Zora Tung, Panupong Pasupat, and Ming{-}Wei Chang.
  2020.
\newblock \href {http://arxiv.org/abs/2002.08909} {{REALM:} retrieval-augmented
  language model pre-training}.
\newblock \emph{CoRR}, abs/2002.08909.

\bibitem[{Hou et~al.(2020)Hou, Huang, Shang, Jiang, Chen, and Liu}]{wnli1}
Lu~Hou, Zhiqi Huang, Lifeng Shang, Xin Jiang, Xiao Chen, and Qun Liu. 2020.
\newblock \href
  {https://proceedings.neurips.cc/paper/2020/file/6f5216f8d89b086c18298e043bfe48ed-Paper.pdf}
  {Dynabert: Dynamic bert with adaptive width and depth}.
\newblock In \emph{Advances in Neural Information Processing Systems},
  volume~33, pages 9782--9793. Curran Associates, Inc.

\bibitem[{Houlsby et~al.(2019)Houlsby, Giurgiu, Jastrzebski, Morrone,
  de~Laroussilhe, Gesmundo, Attariyan, and Gelly}]{bettertune2}
Neil Houlsby, Andrei Giurgiu, Stanislaw Jastrzebski, Bruna Morrone, Quentin
  de~Laroussilhe, Andrea Gesmundo, Mona Attariyan, and Sylvain Gelly. 2019.
\newblock Parameter-efficient transfer learning for nlp.
\newblock In \emph{International Conference on Machine Learning}, pages
  2790--2799.

\bibitem[{Howard and Ruder(2018)}]{finetune3}
Jeremy Howard and Sebastian Ruder. 2018.
\newblock \href {https://doi.org/10.18653/v1/P18-1031} {Universal language
  model fine-tuning for text classification}.
\newblock In \emph{Proceedings of the 56th Annual Meeting of the Association
  for Computational Linguistics (Volume 1: Long Papers)}, pages 328--339,
  Melbourne, Australia. Association for Computational Linguistics.

\bibitem[{Huang et~al.(2021)Huang, Hou, Shang, Jiang, Chen, and Liu}]{wnli3}
Zhiqi Huang, Lu~Hou, Lifeng Shang, Xin Jiang, Xiao Chen, and Qun Liu. 2021.
\newblock \href {https://doi.org/10.18653/v1/2021.acl-long.509} {{G}host{BERT}:
  Generate more features with cheap operations for {BERT}}.
\newblock In \emph{Proceedings of the 59th Annual Meeting of the Association
  for Computational Linguistics and the 11th International Joint Conference on
  Natural Language Processing (Volume 1: Long Papers)}, pages 6512--6523,
  Online. Association for Computational Linguistics.

\bibitem[{Jiang et~al.(2020)Jiang, He, Chen, Liu, Gao, and Zhao}]{smart}
Haoming Jiang, Pengcheng He, Weizhu Chen, Xiaodong Liu, Jianfeng Gao, and Tuo
  Zhao. 2020.
\newblock \href {https://doi.org/10.18653/v1/2020.acl-main.197} {{SMART}:
  Robust and efficient fine-tuning for pre-trained natural language models
  through principled regularized optimization}.
\newblock In \emph{Proceedings of the 58th Annual Meeting of the Association
  for Computational Linguistics}, pages 2177--2190, Online. Association for
  Computational Linguistics.

\bibitem[{Kingma and Ba(2014)}]{Adamax}
Diederik~P. Kingma and Jimmy Ba. 2014.
\newblock \href {https://doi.org/10.48550/ARXIV.1412.6980} {Adam: A method for
  stochastic optimization}.
\newblock \emph{arXiv preprint arXiv:1412.6980}.

\bibitem[{Klinker(2010)}]{exp-average1}
Frank Klinker. 2010.
\newblock \href {https://doi.org/10.1007/s00591-010-0080-8} {Exponential moving
  average versus moving~exponential~average}.
\newblock \emph{Mathematische Semesterberichte}, 58(1):97--107.

\bibitem[{Lee et~al.(2020)Lee, Cho, and Kang}]{mixout}
Cheolhyoung Lee, Kyunghyun Cho, and Wanmo Kang. 2020.
\newblock \href {https://openreview.net/forum?id=HkgaETNtDB} {Mixout: Effective
  regularization to finetune large-scale pretrained language models}.
\newblock In \emph{International Conference on Learning Representations}.

\bibitem[{Lee et~al.(2019)Lee, Ajanthan, and Torr}]{sensitivity7}
Namhoon Lee, Thalaiyasingam Ajanthan, and Philip Torr. 2019.
\newblock \href {https://openreview.net/forum?id=B1VZqjAcYX} {{SNIP}:
  {Single}-{Shot} {Network} {Pruning} {Based} {on} {Connectiono}
  {Sensitivity}}.
\newblock In \emph{International Conference on Learning Representations}.

\bibitem[{Levesque et~al.(2012)Levesque, Davis, and Morgenstern}]{WNLI}
Hector Levesque, Ernest Davis, and Leora Morgenstern. 2012.
\newblock The winograd schema challenge.
\newblock In \emph{Thirteenth International Conference on the Principles of
  Knowledge Representation and Reasoning}.

\bibitem[{Liang et~al.(2022)Liang, Jiang, Zuo, He, Liu, Gao, Chen, and
  Zhao}]{sage}
Chen Liang, Haoming Jiang, Simiao Zuo, Pengcheng He, Xiaodong Liu, Jianfeng
  Gao, Weizhu Chen, and Tuo Zhao. 2022.
\newblock \href {https://openreview.net/forum?id=cuvga_CiVND} {No parameters
  left behind: Sensitivity guided adaptive learning rate for training large
  transformer models}.
\newblock In \emph{International Conference on Learning Representations}.

\bibitem[{Liu(2019)}]{bertnb2}
Yang Liu. 2019.
\newblock \href {http://arxiv.org/abs/1903.10318} {Fine-tune {BERT} for
  extractive summarization}.
\newblock \emph{CoRR}, abs/1903.10318.

\bibitem[{Liu et~al.(2019)Liu, Ott, Goyal, Du, Joshi, Chen, Levy, Lewis,
  Zettlemoyer, and Stoyanov}]{RoBERTa}
Yinhan Liu, Myle Ott, Naman Goyal, Jingfei Du, Mandar Joshi, Danqi Chen, Omer
  Levy, Mike Lewis, Luke Zettlemoyer, and Veselin Stoyanov. 2019.
\newblock Roberta: A robustly optimized bert pretraining approach.
\newblock \emph{arXiv preprint arXiv:1907.11692}.

\bibitem[{Molchanov et~al.(2019)Molchanov, Mallya, Tyree, Frosio, and
  Kautz}]{sensitivity4}
Pavlo Molchanov, Arun Mallya, Stephen Tyree, Iuri Frosio, and Jan Kautz. 2019.
\newblock \href {https://doi.org/10.1109/CVPR.2019.01152} {Importance
  estimation for neural network pruning}.
\newblock In \emph{2019 IEEE/CVF Conference on Computer Vision and Pattern
  Recognition (CVPR)}, pages 11256--11264.

\bibitem[{Molchanov et~al.(2017)Molchanov, Tyree, Karras, Aila, and
  Kautz}]{sensitivity3}
Pavlo Molchanov, Stephen Tyree, Tero Karras, Timo Aila, and Jan Kautz. 2017.
\newblock \href {https://openreview.net/forum?id=SJGCiw5gl} {Pruning
  convolutional neural networks for resource efficient inference}.
\newblock In \emph{5th International Conference on Learning Representations,
  {ICLR} 2017, Toulon, France, April 24-26, 2017, Conference Track
  Proceedings}. OpenReview.net.

\bibitem[{Mosbach et~al.(2020)Mosbach, Andriushchenko, and
  Klakow}]{bettertune3}
Marius Mosbach, Maksym Andriushchenko, and Dietrich Klakow. 2020.
\newblock \href {https://doi.org/10.48550/ARXIV.2006.04884} {On the stability
  of fine-tuning bert: Misconceptions, explanations, and strong baselines}.
\newblock \emph{arXiv preprint arXiv:2006.04884}.

\bibitem[{Peters et~al.(2019)Peters, Ruder, and Smith}]{bettertune1}
Matthew~E. Peters, Sebastian Ruder, and Noah~A. Smith. 2019.
\newblock \href {https://doi.org/10.18653/v1/W19-4302} {To tune or not to tune?
  adapting pretrained representations to diverse tasks}.
\newblock In \emph{Proceedings of the 4th Workshop on Representation Learning
  for NLP (RepL4NLP-2019)}, pages 7--14, Florence, Italy. Association for
  Computational Linguistics.

\bibitem[{Qiu et~al.(2020)Qiu, Sun, Xu, Shao, Dai, and Huang}]{bertnb4}
XiPeng Qiu, TianXiang Sun, YiGe Xu, YunFan Shao, Ning Dai, and XuanJing Huang.
  2020.
\newblock \href {https://doi.org/10.1007\%2Fs11431-020-1647-3} {Pre-trained
  models for natural language processing: A survey}.
\newblock \emph{Science China Technological Sciences}, 63(10):1872--1897.

\bibitem[{Raffel et~al.(2019)Raffel, Shazeer, Roberts, Lee, Narang, Matena,
  Zhou, Li, and Liu}]{bad2}
Colin Raffel, Noam Shazeer, Adam Roberts, Katherine Lee, Sharan Narang, Michael
  Matena, Yanqi Zhou, Wei Li, and Peter~J. Liu. 2019.
\newblock \href {https://doi.org/10.48550/ARXIV.1910.10683} {Exploring the
  limits of transfer learning with a unified text-to-text transformer}.
\newblock \emph{arXiv preprint arXiv:1910.10683}.

\bibitem[{Rajpurkar et~al.(2016)Rajpurkar, Zhang, Lopyrev, and Liang}]{QNLI}
Pranav Rajpurkar, Jian Zhang, Konstantin Lopyrev, and Percy Liang. 2016.
\newblock \href {https://doi.org/10.18653/v1/D16-1264} {{SQ}u{AD}: 100,000+
  questions for machine comprehension of text}.
\newblock In \emph{Proceedings of the 2016 Conference on Empirical Methods in
  Natural Language Processing}, pages 2383--2392, Austin, Texas. Association
  for Computational Linguistics.

\bibitem[{Sanh et~al.(2020)Sanh, Wolf, and Rush}]{redundancy2}
Victor Sanh, Thomas Wolf, and Alexander Rush. 2020.
\newblock \href
  {https://proceedings.neurips.cc/paper/2020/file/eae15aabaa768ae4a5993a8a4f4fa6e4-Paper.pdf}
  {Movement pruning: Adaptive sparsity by fine-tuning}.
\newblock In \emph{Advances in Neural Information Processing Systems},
  volume~33, pages 20378--20389. Curran Associates, Inc.

\bibitem[{Shahidi et~al.(2020)Shahidi, Li, and Lin}]{exp-average3}
Hamidreza Shahidi, Ming Li, and Jimmy Lin. 2020.
\newblock \href {https://doi.org/10.18653/v1/2020.acl-main.355} {Two birds, one
  stone: A simple, unified model for text generation from structured and
  unstructured data}.
\newblock In \emph{Proceedings of the 58th Annual Meeting of the Association
  for Computational Linguistics}, pages 3864--3870, Online. Association for
  Computational Linguistics.

\bibitem[{Sun et~al.(2019)Sun, Qiu, Xu, and Huang}]{finetune2}
Chi Sun, Xipeng Qiu, Yige Xu, and Xuanjing Huang. 2019.
\newblock How to fine-tune bert for text classification?
\newblock In \emph{Chinese Computational Linguistics}, pages 194--206, Cham.
  Springer International Publishing.

\bibitem[{Wadden et~al.(2019)Wadden, Wennberg, Luan, and
  Hajishirzi}]{finetune1}
David Wadden, Ulme Wennberg, Yi~Luan, and Hannaneh Hajishirzi. 2019.
\newblock \href {https://aclanthology.org/D19-1585} {Entity, relation, and
  event extraction with contextualized span representations}.
\newblock In \emph{Proceedings of the 2019 Conference on Empirical Methods in
  Natural Language Processing and the 9th International Joint Conference on
  Natural Language Processing (EMNLP-IJCNLP)}, pages 5784--5789, Hong Kong,
  China. Association for Computational Linguistics.

\bibitem[{Wang et~al.(2018)Wang, Singh, Michael, Hill, Levy, and Bowman}]{glue}
Alex Wang, Amanpreet Singh, Julian Michael, Felix Hill, Omer Levy, and Samuel
  Bowman. 2018.
\newblock \href {https://doi.org/10.18653/v1/W18-5446} {{GLUE}: A multi-task
  benchmark and analysis platform for natural language understanding}.
\newblock In \emph{Proceedings of the 2018 {EMNLP} Workshop {B}lackbox{NLP}:
  Analyzing and Interpreting Neural Networks for {NLP}}, pages 353--355,
  Brussels, Belgium. Association for Computational Linguistics.

\bibitem[{Warstadt et~al.(2019)Warstadt, Singh, and Bowman}]{CoLA}
Alex Warstadt, Amanpreet Singh, and Samuel~R. Bowman. 2019.
\newblock \href {https://doi.org/10.1162/tacl_a_00290} {Neural network
  acceptability judgments}.
\newblock \emph{Transactions of the Association for Computational Linguistics},
  7:625--641.

\bibitem[{Williams et~al.(2018)Williams, Nangia, and Bowman}]{MNLI1}
Adina Williams, Nikita Nangia, and Samuel Bowman. 2018.
\newblock \href {http://aclweb.org/anthology/N18-1101} {A broad-coverage
  challenge corpus for sentence understanding through inference}.
\newblock In \emph{Proceedings of the 2018 Conference of the North American
  Chapter of the Association for Computational Linguistics: Human Language
  Technologies, Volume 1 (Long Papers)}, pages 1112--1122. Association for
  Computational Linguistics.

\bibitem[{Wu et~al.(2022)Wu, Wu, Qi, and Huang}]{noisytune}
Chuhan Wu, Fangzhao Wu, Tao Qi, and Yongfeng Huang. 2022.
\newblock \href {https://doi.org/10.18653/v1/2022.acl-short.76} {{N}oisy{T}une:
  A little noise can help you finetune pretrained language models better}.
\newblock In \emph{Proceedings of the 60th Annual Meeting of the Association
  for Computational Linguistics (Volume 2: Short Papers)}, pages 680--685,
  Dublin, Ireland. Association for Computational Linguistics.

\bibitem[{Xiao et~al.(2019)Xiao, Wang, and Rajasekaran}]{sensitivity6}
Xia Xiao, Zigeng Wang, and Sanguthevar Rajasekaran. 2019.
\newblock \href
  {https://proceedings.neurips.cc/paper/2019/file/4efc9e02abdab6b6166251918570a307-Paper.pdf}
  {Autoprune: Automatic network pruning by regularizing auxiliary parameters}.
\newblock In \emph{Advances in Neural Information Processing Systems},
  volume~32. Curran Associates, Inc.

\bibitem[{Zhu et~al.(2020{\natexlab{a}})Zhu, Cheng, Gan, Sun, Goldstein, and
  Liu}]{FreeLB}
Chen Zhu, Yu~Cheng, Zhe Gan, Siqi Sun, Tom Goldstein, and Jingjing Liu.
  2020{\natexlab{a}}.
\newblock \href {https://openreview.net/forum?id=BygzbyHFvB} {Freelb: Enhanced
  adversarial training for natural language understanding}.
\newblock In \emph{International Conference on Learning Representations}.

\bibitem[{Zhu et~al.(2020{\natexlab{b}})Zhu, Xia, Wu, He, Qin, Zhou, Li, and
  Liu}]{bertnb3}
Jinhua Zhu, Yingce Xia, Lijun Wu, Di~He, Tao Qin, Wengang Zhou, Houqiang Li,
  and Tie-Yan Liu. 2020{\natexlab{b}}.
\newblock \href {https://arxiv.org/abs/2002.06823} {Incorporating bert into
  neural machine translation}.
\newblock \emph{arXiv preprint arXiv:2002.06823}.

\bibitem[{Zhuang et~al.(2022)Zhuang, Wen, and Zhang}]{exp-average2}
Weiming Zhuang, Yonggang Wen, and Shuai Zhang. 2022.
\newblock \href {https://openreview.net/forum?id=oVE1z8NlNe} {Divergence-aware
  federated self-supervised learning}.
\newblock In \emph{International Conference on Learning Representations}.

\end{thebibliography}
\bibliographystyle{acl_natbib}

\begin{table*}[ht]
\centering
\begin{tabular}{lcccccccc}
\Xhline{1.6pt}
\textbf{Model}  & \textbf{COLA} & \textbf{MRPC} & \textbf{RTE} & \textbf{STS-B}  & \textbf{QQP} & \textbf{QNLI} & \textbf{MNLI} & \textbf{SST} \\
\hline
$\mbox{BERT}_{\mbox{base}}$  & 1e-4  & 1e-4  & 1e-4 & 2e-4 & 1e-4 & 2e-4 & 8e-5 & 8e-5 \\
\hline
$\mbox{RoBERTa}_{\mbox{large}}$  & 3e-5  & 5e-5  & 5e-5 & 5e-5 & 1e-4 & 1e-5 & 3e-5 & 3e-5 \\
\hline
$\mbox{BERT}_{\mbox{base}}$+PATS  & 1e-4  & 3e-4  & 3e-4 & 3e-4 & 2e-4 & 2e-4 & 1e-4 & 3e-4 \\
\hline
$\mbox{RoBERTa}_{\mbox{large}}$+PATS  & 8e-5  & 8e-5  & 8e-5 & 5e-5 & 1e-4 & 3e-5 & 1e-5 & 1e-5 \\
\Xhline{1.6pt}
\end{tabular}

\caption{Learning rate settings for PATS on the tasks of the GLUE benchmark.}
\label{tab:lr}
\end{table*}

\begin{table*}[!ht]
\centering
\begin{tabular}{c|c}
    \Xhline{1.6pt}
    Hyperparameters & Range \\
    \hline
    $\lambda$ & \{5e-7, 8e-7, 1e-6, 2e-6, 3e-6\}\\
    $\gamma$ & \{1e-3, 2e-3, 3e-3, 5e-3, 8e-3,2e-2\}\\
    $\beta$ & \{0.5, 0.55, 0.6, 0.65, 0.7, 0.75, 0.8, 0.85\}\\
    learning rate & \{1e-5, 3e-5, 5e-5, 7e-5, 8e-5, 1e-4, 2e-4, 3e-4, 5e-4\}\\
    \Xhline{1.6pt}
\end{tabular}
\caption{Searching ranges of hyperparameters in our experiments.}
\label{tab:hyperparameters}
\end{table*}
\appendix

\section{Appendix}

\subsection{Datasets}
\label{sec:datasets}
The experiments are conducted on the GLUE benchmark, which contains several types of Natural Language Understanding (NLU) tasks such as linguistic acceptability (CoLA, \citealt{CoLA}), text similarity (STS-B, \citealt{STS-b}) and natural language inference (RTE \& MNLI \& QNLI, \citealt{RTE1, RTE2, RTE3, RTE4, MNLI1, MNLI2, QNLI}) tasks. Among the nine tasks, WNLI \citep{WNLI} task is excluded in our experiments, on which BERT-like models have no obvious advantage over other mainstream baselines \citep{wnli1,wnli2,wnli3}. Consistent with previous works \citep{dev, noisytune}, we evaluate results on the dev set of GLUE.


\subsection{Training Details}
\label{sec:appendix}
For all the baseline models and our proposed PATS, we adopt a linear-decay learning rate schedule and choose Adamax \citep{Adamax} which is the best-performing optimizer for baseline models on the GLUE benchmark to optimize the training. In PATS, we perturb the parameters of all the encoder layers except the Layer Normalization layers. 
In our training process, we set $\lambda=2\times10^{-6}$, $\gamma=0.002$, $\beta=0.75$, $p=0.2$ for all tasks. In addition, we adopt a linear warm-up learning rate schedule with 0.1 of total training iterations. The batch size of models is uniformly set to 32. We post the best performance models on each task after 10 epochs of training. The learning rates that yields the best generalization performance of models optimized by PATS and Standard PLM fine-tuning on each task are listed in Table \ref{tab:lr}. We present the searching range of hyperparameters in Table \ref{tab:hyperparameters}.

Our implementation is based on the MT-DNN code-base.\footnote{\href{https://github.com/namisan/mt-dnn}{https://github.com/namisan/mt-dnn}}. And we use Nvidia V100 GPUs for all experiments.

\subsection{Other Implemention Details}
\label{ap:al}
All datasets of the GLUE benchmark are downloaded from \href{https:// gluebenchmark.com/tasks}{https:// gluebenchmark.com/tasks}. For the baseline model SAGE, we use the code from the Github respository \href{https://github.com/cliang1453/SAGE}{https://github.com/cliang1453/SAGE}. The other baseline models are implemented by ourselves.

For the distribution of sensitivity shown in Fig. \ref{fig:distribution}, we discard some outliers and only choose the parameters with sensitivity in the range of [5e-8, 1e-5] for visualization.

\end{document}